\documentclass{article}

\PassOptionsToPackage{numbers}{natbib}
\usepackage[final]{bdl_2018}

\usepackage[utf8]{inputenc} % allow utf-8 input
\usepackage[T1]{fontenc}    % use 8-bit T1 fonts
\usepackage{hyperref}       % hyperlinks
\usepackage{url}            % simple URL typesetting
\usepackage{booktabs}       % professional-quality tables
\usepackage{amsfonts}       % blackboard math symbols
\usepackage{amsmath}
\usepackage{mathtools}
\usepackage{caption}
\usepackage[pdftex]{graphicx}
\usepackage{nicefrac}       % compact symbols for 1/2, etc.
\usepackage{microtype}      % microtypography

\title{Variational Dropout via Empirical Bayes}

\author{
  Valery~Kharitonov ${}^{1}$\\
  \texttt{kharvd@gmail.com} \\
  \And
  Dmitry~Molchanov ${}^{1,2}$\\
  \texttt{dmolch111@gmail.com}
  \And
  Dmitry~Vetrov ${}^{1,2}$\\
  \texttt{vetrovd@yandex.ru}
  \And
  \normalfont ${}^1$National Research University Higher School of Economics, Joint Samsung-HSE Lab\\
  ${}^2$Samsung AI Center in Moscow
}

\newcommand{\Ee}[2][\null]{\mathbb{E}_{#1}\left[{#2}\right]}
\newcommand{\KL}[2]{\mathrm{D}_{KL}\!\left({#1} \, || \, {#2}\right)}

\begin{document}

\maketitle
\vspace{-1.2em}
\begin{abstract}
  We study the Automatic Relevance Determination procedure applied to deep neural networks. 
  We show that ARD applied to Bayesian DNNs with Gaussian approximate posterior distributions
  leads to a variational bound similar to that of variational dropout, and in
  the case of a fixed dropout rate, objectives are exactly the same. 
  Experimental results show that the two approaches yield comparable results in practice even when
  the dropout rates are trained. This leads to an alternative Bayesian interpretation of 
  dropout and mitigates some of the theoretical issues that arise with the use 
  of improper priors in the variational dropout model. Additionally, we explore the use of the hierarchical priors in ARD and show that it helps achieve higher sparsity for the same accuracy.
\end{abstract}
\vspace{-1.2em}
\section{Introduction}
Dropout \cite{srivastava2014} is a popular regularization method for neural networks that can be interpreted as a form of approximate Bayesian inference \cite{kingma2015,gal2016}. Sparse variational dropout (Sparse VD) \cite{molchanov2017} further extends this approach and shows that it can be used to significantly prune neural networks. However, it was recently argued \cite{hron2018} that the use of an improper prior distribution in current formulations of variational dropout leads to an improper posterior, and as such, this model cannot be used to provide a principled interpretation of the empirical behavior of the dropout procedure. In this paper, we use variational inference to perform automatic relevance determination procedure \cite{mackay1995,neal1996} to arrive at an alternative interpretation of dropout that does not have such a drawback. The derived objective is remarkably similar to the approximation in Sparse VD and empirical observations confirm that the two models are effectively equivalent.

\section{Variational Automatic Relevance Determination for neural networks}
Let $\mathcal{D} = (x_i, y_i)_{i=1}^N$ be a dataset of $N$ samples where $x_i$ are observable variables and $y_i$ are the targets. Suppose that we have some parametric model $p(\mathcal{D}|w) = \prod_{i=1}^N p(y_i | x_i, w)$, $w \in \mathbb{R}^D$ (e.g. a deep neural network) mapping $x$ to the corresponding $y$ using parameters $w$. The parameters $w$ have a prior distribution $p(w | \tau)$ which is itself parameterized by hyperparameters $\tau \in \mathbb{R}^H$. Following the Bayesian approach, we wish to find the posterior distribution $p(w | \mathcal{D}, \tau) = p(\mathcal{D} | w) p(w|\tau) / p(\mathcal{D} | \tau)$. We choose hyperparameters $\tau$ such that the marginal likelihood (evidence) of the dataset is maximized:
\[ \tau^* = \arg \max_{\tau}\, p(\mathcal{D} | \tau) = \arg \max_{\tau}\, \int p(\mathcal{D} | w) p(w|\tau) \, \mathrm{dw}.\]
This is the \emph{empirical Bayes} (EB) approach to hyperparameter selection. In particular, when $p(w|\tau) = \prod_{i=1}^D \mathcal{N}(w_i \,|\, 0, \tau_i^{-1})$, this procedure is called \emph{automatic relevance determination} (ARD) \cite{mackay1995,neal1996}.

% Relevance vector machine~(RVM)~\cite{tipping2000} and Soft Weight Shairng~\cite{ullrich2017} are examples of the EB used for automatic relevance determination.

Since in the case of deep neural networks the marginal likelihood is intractable, we use doubly stochastic variational inference (DSVI) \cite{titsias2014} to find an approximate posterior $q(w | \phi)$ from some parametric family. This is achieved by maximizing the \emph{evidence lower bound (ELBO)} w.r.t. variational parameters $\phi$ and hyperparameters $\tau$:
\[ \log p(\mathcal{D} | \tau) \geq \mathcal{L}(\phi, \tau) = \Ee[q(w|\phi)]{\log p(\mathcal{D}|w)} - \KL{q(w|\phi)}{p(w|\tau)} \to \max_{\phi, \tau}.\]
Now suppose that $p(w|\tau) = \prod_{i=1}^D \mathcal{N}(w_i | 0, \tau_i^{-1})$ (ARD prior) and $q(w|\mu, \sigma) = \prod_{i=1}^D \mathcal{N}(w_i | \mu_i, \sigma_i^2)$.
The optimal value for hyperparameters $\tau$ in this case can be found analytically \cite{titsias2014}: $\tau_i^* = (\mu_i^2 + \sigma_i^2)^{-1}$. The ELBO then takes the following form \cite{titsias2014}:
\begin{align}\label{eq:ard_elbo}
\mathcal{L}_{ARD}(\mu, \sigma) &= \sum_{i=1}^N \Ee[q(w|\mu, \sigma)]{\log p(y_i|x_i, w)} - \frac{1}{2} \sum_{j=1}^D \log \left(1 + \frac{\mu_j^2}{\sigma_j^2} \right) \\
&= \mathcal{L}_{\mathcal{D}}(\mu, \sigma) + \mathcal{R}_{ARD}(\mu, \sigma) \to \max_{\mu, \sigma}. \nonumber
\end{align}
In practice, we estimate the gradients of the ELBO w.r.t. the variational parameters using the local reparameterization trick \cite{kingma2015}. The objective \eqref{eq:ard_elbo} was first derived in the context of linear models \cite{titsias2014}.
% It has two terms. The first term is the likelihood of the data. The second term can be interpreted as a regularizer. It encourages sparse solutions: the KL divergence is minimized either when $\mu_i$ is zero or $\sigma_i$ is large. When both $\mu_i$ and $\sigma_i$ are small, that means that both the prior and the posterior for the corresponding $w_i$ are concentrated at zero and the weight can be removed from the network. 

\section{Connection with variational dropout}
Variational dropout (VD) \cite{kingma2015} is a generalization of Gaussian dropout which interprets it as an approximate Bayesian inference procedure. It puts an improper scale-invariant log-uniform prior on the weights of a neural network with fully-connected and convolutional layers and uses a factorized Gaussian approximation to the true posterior.

Let us now consider a restricted variational approximation $q(w|\mu) = \prod_{i=1}^D \mathcal{N}(w_i\,|\,\mu_i, \alpha \mu_i^2)$ for some constant $\alpha > 0$ in the ARD objective (\ref{eq:ard_elbo}). The regularizer term is now constant and does not affect optimization, so the objective takes the form:
\begin{equation}\label{eq:dropout_elbo}
\widetilde{\mathcal{L}}_{ARD}(\mu) = \mathcal{L}_{\mathcal{D}}(\mu) = \sum_{i=1}^N \Ee[q(w|\mu)]{\log p(y_i|x_i, w)} \to \max_{\mu}.
\end{equation}
It can be shown \cite{kingma2015} that optimizing such functional is equivalent to training a neural network with Gaussian dropout which puts multiplicative normal noise $\mathcal{N}(1, \alpha)$ on the input of each dense and convolutional layer in the net, while ignoring the dependencies between the output units of the layer. Note that unlike variational dropout \cite{kingma2015}, we did not have to use an improper prior to arrive at this objective.

We can assign individual dropout rates to each weight in the network and optimize with respect to them. Such an approach \cite{molchanov2017} leads to very sparse solutions where most of the weights of a network are assigned high dropout rates, thus effectively being pruned from the network. To tune individual dropout rates in the ARD model, we set $q(w|\mu, \alpha) = \prod_{i=1}^D \mathcal{N}(w_i\,|\,\mu_i, \alpha_i \mu_i^2)$ which yields the following objective (ARD Dropout):
\begin{equation}\label{eq:multiplicative_dropout_elbo}
\mathcal{L}_{ARD}(\mu, \alpha) = \mathcal{L}_{\mathcal{D}}(\mu, \alpha) - \frac{1}{2} \sum_{j=1}^D \log (1 + \alpha_j^{-1}) \to \max_{\mu, \alpha}.
\end{equation}

Compare this to the approximation for the Sparse VD objective  \cite{molchanov2017}:
\begin{gather}\label{eq:multiplicative_vardrop_elbo}
\mathcal{L}_{SVDO}(\mu, \alpha) \approx \mathcal{L}_{\mathcal{D}}(\mu, \alpha) + \sum_{j=1}^D \left[ k_1 \sigma(k_2 + k_3 \log \alpha_j) - \frac{1}{2} \log (1 + \alpha_j^{-1}) + C \right] \to \max_{\mu, \alpha}, \\ \nonumber
\end{gather}
where $k_1 = 0.63576, \; k_2 = 1.87320,  \; k_3 = 1.48695, \; C = -k_1.$ As we can see, both \eqref{eq:multiplicative_dropout_elbo} and \eqref{eq:multiplicative_vardrop_elbo} have the same $\log$ terms in the regularizer and the sigmoid term in $\mathcal{L}_{SVDO}(\mu, \alpha)$ is bounded. It is also easy to show that Sparse VD objective is actually a lower bound on the ARD objective. Furthermore, when $\alpha_j \to \infty$, $\mathcal{L}_{SVDO}(\mu, \alpha)$ approaches $\mathcal{L}_{ARD}(\mu, \alpha)$ from below. The common term in the objectives is the one that encourages sparsity, so we expect that ARD would perform similarly to Sparse VD in terms of compression rate. In Table \ref{tab:results}, we report accuracy and compression for networks trained with ARD and Sparse VD objectives (see Appendix \ref{app:experiments} for details). As we can see, both models show comparable sparsity while maintaining low classification error rate.

The proposed ARD dropout interpretation mitigates one of the issues described in Hron~et~al.~(2018) \cite{hron2018}. The paper identifies two problems with variational dropout: (a) the use of improper prior distribution and (b) singularity of the approximate posterior distribution. Now that we do not use the log-uniform prior anymore, our ARD dropout model is defined correctly, and both the prior and the true posterior distributions for each value of $\tau$ are now proper, which fixes (a), and (b) is only present in the model with correlated weight noise which is not the case for the ARD model.

\section{Accuracy-compression trade-off using Gamma hyperprior}
We can go further and introduce a hyperprior over hyperparameters $\tau$ and then perform \emph{maximum a posteriori} estimation for them. Suppose that $p(\tau_i | a, b) = \mathrm{Gamma}(\tau_i | a, b)$. If we maximize the evidence w.r.t. $\tau$ in such model (so-called MAP-II estimation), it leads to the following objective when $a > 1/2$, $b > 0$ (see Appendix \ref{app:gamma} for details):
\[ \mathcal{L}_\Gamma(\mu, \sigma) = \mathcal{L}_{\mathcal{D}}(\mu, \sigma) + \sum_{j=1}^D \left[ \frac{1}{2} \log \frac{\sigma_j^2}{(\sigma_j^2 + \mu_j^2 + 2b)^{2a - 1}} + C \right] \to \max_{\mu,\sigma},\]
where $C = 1 - a + a \log b - \log \Gamma(a) + \frac{1}{2} (2a - 1) \log (2a - 1), $ and $\Gamma$ is the gamma function. Unlike ARD or Sparse VD models, we now have tunable parameters $a, b$ of the hyperprior. With $a = 1$, $b = 0$ (which gives an improper prior) and ignoring the constant term, we get the usual ARD objective.  Smaller values of $a$ (around 0.5) reduce pruning without rescaling the regularizer term, which is a common way to prevent underfitting \cite{louizos2017,ullrich2017}. Moreover, this approach helps achieve higher values of sparsity for the same accuracy level. See Table \ref{tab:results_gamma} for empirical results for different values of $a$.

Interestingly, if we approach this model in a fully Bayesian way and marginalize the hyperparameters out, we obtain the marginal prior $p(w)$ distributed according to the generalized Student t-distribution $ p(w_i) = \mathrm{Student}(\nu = 2a, \mu = 0, \lambda = a/b)$.
In particular, the log-uniform prior used in Sparse VD can be considered a limiting case of this model when $a = b \to 0$. See Appendix \ref{app:student} for details.

\begin{table}
\centering
\begin{minipage}[t]{0.45\linewidth}
  \centering
  \small
  \begin{tabular}{lll}
    \toprule
    Objective  & Error \% & Compression \\
    \midrule
    \multicolumn{3}{c}{MNIST (LeNet-5)}                   \\
    Sparse VD  & $0.81$ ($\pm 0.04$) & $136$ ($\pm 8$)  \\
    \textbf{ARD Dropout} & $0.76$ ($\pm 0.09$) & $132$ ($\pm 13$)          \\
    \midrule
    \multicolumn{3}{c}{CIFAR-10 (VGG-like)}                   \\
    Sparse VD  & $8.08$ ($\pm 0.27$) & $38$ ($\pm 1$)          \\
    \textbf{ARD Dropout}  & $7.75$ ($\pm 0.28$) & $34$ ($\pm 1$)          \\
    \bottomrule
  \end{tabular}
  \vspace{1em}
  \caption{\small Empirical comparison of ARD and sparse variational dropout. Compression is defined as the total number of weights in the network divided by the number of non-zero weights after trimming.}
  \label{tab:results}
\end{minipage}
\hspace{1em}
\begin{minipage}[t]{0.5\linewidth}
  % \includegraphics[width=\linewidth]{dkl_plot.pdf}
% \vspace{0.5em}
% \captionof{figure}{\small Plots of regularizer terms in variational dropout and ARD objectives for a single weight as a function of $\log \alpha_j$.}
% \label{fig:dkl_plot}
  \centering
  \small
  \begin{tabular}{lll}
    \toprule
    $a$  & Error \% & Compression \\
    \midrule
    \multicolumn{3}{c}{CIFAR-10 (VGG-like)}                   \\
    0.505  & $7.46$ ($\pm 0.24$) & $52$ ($\pm 1$)          \\
    0.510  & $7.65$ ($\pm 0.25$) & $85$ ($\pm 1$)          \\
    0.515  & $7.88$ ($\pm 0.13$) & $114$ ($\pm 1$)          \\
    0.520  & $7.94$ ($\pm 0.18$) & $136$ ($\pm 2$)          \\
    \bottomrule
  \end{tabular}
  \vspace{1em}
  \caption{\small Classification error and compression for Gamma hyperprior MAP-II model, where $b = 10^{-8}$.}
  \label{tab:results_gamma}
\end{minipage}
\vspace{-1.5em}
\end{table}

\section{Conclusion}
We have shown that it is possible to overcome the theoretical difficulties with variational dropout using a variational approximation to the ARD procedure with fully factorized Gaussian variational posterior distributions. It does not require the use of improper priors and leads to a similar variational lower bound, and thus can be seen as an alternative to variational dropout. Our theoretical study is supported by experimental results that show that in practice, both methods achieve comparably high values of sparsity without a significant drop in classification accuracy. Additionally, such an approach allows applying a hierarchical prior to the hyperparameters, which gives an option to trade off between accuracy and sparsity.

\subsubsection*{Acknowledgements}
We would like to thank Dmitry Kropotov, Arsenii Ashukha, Kirill Neklyudov, and Dmitrii Podoprikhin for valuable discussions and feedback.

\bibliography{nips_2018}{}

% [1] M. Titsias and M. Lázaro-Gredilla, “Doubly Stochastic Variational Bayes for non-Conjugate Inference,” Proc. 31st Int. Conf. Mach. Learn., vol. 32, pp. 1971–1979, 2014.

% [1] Alexander, J.A.\ \& Mozer, M.C.\ (1995) Template-based algorithms
% for connectionist rule extraction. In G.\ Tesauro, D.S.\ Touretzky and
% T.K.\ Leen (eds.), {\it Advances in Neural Information Processing
%   Systems 7}, pp.\ 609--616. Cambridge, MA: MIT Press.

% [2] Bower, J.M.\ \& Beeman, D.\ (1995) {\it The Book of GENESIS:
%   Exploring Realistic Neural Models with the GEneral NEural SImulation
%   System.}  New York: TELOS/Springer--Verlag.

% [3] Hasselmo, M.E., Schnell, E.\ \& Barkai, E.\ (1995) Dynamics of
% learning and recall at excitatory recurrent synapses and cholinergic
% modulation in rat hippocampal region CA3. {\it Journal of
%   Neuroscience} {\bf 15}(7):5249-5262.

\appendix

\section{Experimental setting}
\label{app:experiments}
We compare ARD dropout with Sparse VD \cite{molchanov2017}. We train  LeNet-5-Caffe\footnote{\url{https://github.com/BVLC/caffe/blob/master/examples/mnist/lenet.prototxt}} and VGG-like\footnote{\url{http://torch.ch/blog/2015/07/30/cifar.html}} networks on MNIST and CIFAR-10 correspondingly. To reduce the variance of the gradient, we use the additive parameterization \cite{molchanov2017}, training means and logarithms of standard deviations. We train both networks from the same random initialization (with $\log \sigma$ initialized from $\mathcal{N}(-5, 0.1^2)$) for 200 epochs using Adam optimizer with the initial learning rate $10^{-3}$ and minibatch size 100. Starting from epoch 100, we linearly reduce the learning rate to zero. To overcome underfitting, for VGG on CIFAR-10, we scale the regularizer terms in both objectives by $0.05$, and for both networks, we anneal the regularizer term over 20 epochs. For Gamma hyperprior experiments, posterior variances are clipped so that $\log \sigma < -4$ and the regularizer term is not scaled. After training, we evaluate the networks in  ``deterministic mode'': we set the network weights to the means of their approximate posterior distribution. We then trim the weights of the networks setting those with $|\mu_i| < 10^{-2}$ to zero.  We report the mean and the standard deviation for sparsity and error over 5 random seeds.

\section{MAP-II Estimation with Gamma Hyperprior}
\label{app:gamma}
We can introduce a prior distribution over the inverse variance $\tau$:
$$ p(y, w, \tau | x) = p(y| x, w) p(w | \tau) p(\tau), $$
where $p(\tau) = \prod_i \mathrm{Gamma}(\tau_i | a, b)$, and $p(w_i | \tau_i) = \mathcal{N}(w_i \,|\, 0, \tau_i^{-1})$.
Then, we perform maximum a posteriori estimation for the hyperparameters of the prior over the weights. The objective takes the form:
\begin{equation}
\label{eq:gamma_obj}
\mathcal{L}(\mu, \sigma, \tau) = \mathcal{L}_{data}(\mu, \sigma) + \log p(\tau) - \KL{q(w|\mu, \sigma)}{p(w|\tau)} \to \max_{\tau, \mu, \sigma}.
\end{equation}
The KL term is the same as for the regular ARD:
\[ \KL{q(w_i|\mu_i, \sigma_i)}{p(w_i|\tau_i)} = -\frac{1}{2} + \frac{\tau_i (\sigma_i^2 + \mu_i^2)}{2} - \frac{\log (\tau_i \sigma_i^2)}{2}. \]
The regularizer:
\[ \log p(\tau_i) = a \log b - \log \Gamma(a) + (a - 1) \log \tau_i - b \tau_i. \]
Differentiating \eqref{eq:gamma_obj} with respect to $\tau_i$ and equating to zero, we get the optimal value of $\tau_i$ which exists when $a > \frac{1}{2}$:
\[ \tau^*_i = \frac{2a - 1}{\sigma_i^2 + \mu_i^2 + 2b}. \]
Substituting back, we get the following regularizer for a single weight:
\[ R(\mu_i, \sigma_i) = \frac{1}{2} \log \frac{\sigma_i^2}{(\sigma_i^2 + \mu_i^2 + 2b)^{2a - 1}} + C, \]
where
\[ C = 1 - a + a \log b - \log \Gamma(a) + \frac{1}{2} (2a - 1) \log (2a - 1). \]

\section{Full Bayes with Gamma Hyperprior}
\label{app:student}
Instead of a MAP estimation, we can go full Bayes and marginalize over hyperparameters:
\[ p(w_i) = \int p(w_i \,|\, \tau_i) p(\tau_i) \, d \tau_i = 
\int \mathcal{N}(w_i \,|\, 0, \tau_i^{-1}) \cdot \mathrm{Gamma}(\tau_i \,|\, a, b) \, d \tau_i. \]
It is a well-known fact that such Gaussian scale mixture gives  a generalized Student t-distribution:
\[ p(w_i) = \mathrm{Student}(\nu = 2a, \mu = 0, \lambda = a/b). \]
Setting $a = b =\xi$ and taking limit as $\xi$ approaches 0 from above, we get a log-uniform distribution:
\[ p(\log |w_i|) \propto C, \]
which is exactly the prior for variational dropout.

\end{document}